\documentclass[11pt,a4paper]{article}
\usepackage[hyperref]{emnlp-ijcnlp-2019}
\usepackage{times}
\usepackage{latexsym}

\usepackage{url}

\usepackage{diagbox}
\usepackage{multirow}
\usepackage{bm}
\usepackage{graphicx}
\usepackage{amsmath}
\usepackage{amsfonts}
\usepackage{rotating}
\usepackage{tabularx}
\usepackage{subcaption}
\usepackage{xcolor}
\usepackage{pgfplots}

\aclfinalcopy % Uncomment this line for the final submission

\title{Data-Efficient Goal-Oriented Conversation with Dialogue Knowledge Transfer Networks}
\author{
  Igor Shalyminov\textsuperscript{\dag}, Sungjin Lee\textsuperscript{\ddag},
  Arash Eshghi\textsuperscript{\dag}, and Oliver Lemon\textsuperscript{\dag} \\
  \textsuperscript{\dag}Heriot-Watt University, UK \\
  \textsuperscript{\ddag}Amazon Alexa AI, USA \\
  \texttt{\textsuperscript{\dag}\{is33, a.eshghi, o.lemon\}@hw.ac.uk, \textsuperscript{\ddag}sungjinl@amazon.com}
  }

\date{}

\begin{document}
\maketitle

\begin{abstract}
%OL edited version
Goal-oriented dialogue systems are now being widely adopted in industry where it is of key importance to maintain a rapid prototyping cycle for new products and domains. Data-driven dialogue system development has to be adapted to meet this requirement~--- therefore, reducing the amount of data and annotations necessary for training such systems is a central research problem.

In this paper, we present the Dialogue Knowledge Transfer Network (DiKTNet), a state-of-the-art approach to goal-oriented dialogue generation which only uses a few example dialogues (i.e.\ few-shot learning), none of which has to be annotated.
We achieve this by performing a 2-stage training. Firstly, we perform \textit{unsupervised dialogue representation pre-training} on a large source of goal-oriented dialogues in multiple domains, the MetaLWOz corpus. Secondly, at the \textit{transfer stage}, we train DiKTNet using this representation together with 2 other textual knowledge sources with different levels of generality: ELMo encoder and the main dataset's source domains.

Our main dataset is the Stanford Multi-Domain dialogue corpus. We evaluate our model on it in terms of BLEU and Entity F1 scores, and show that our approach significantly and consistently improves upon a series of baseline models as well as over the previous state-of-the-art dialogue generation model, ZSDG. The improvement upon the latter~--- up to \textbf{10\%} in Entity F1 and the average of \textbf{3\%} in BLEU score~--- is achieved using only the equivalent of \textbf{10\%} of  ZSDG's in-domain training data.
% Although the prior work has been presented as a zero-shot model, we show that our approach
\end{abstract}

%original version of abstract:
%Goal-oriented dialogue systems are now approaching the stage of massive adoption in industry. And it's of key importance for industrial environments to maintain rapid prototyping cycle for products and services, e.g. dialogue systems for new products and domains. Data-driven dialogue system development, while being the most flexible and versatile methodology, has to be adapted to meet this requirement. Therefore, developing methods for training such systems with minimal data and annotations is a central problem in dialogue systems research.
%In this paper, we present a state-of-the-art approach to goal-oriented dialogue generation only using several example dialogues (i.e. few-shot learning), none of which has to be annotated.
%We achieve that by performing a 2-stage transfer learning. Firstly, we use we latent dialogue representation trained in an unsupervised way on a greater source of goal-oriented dialogues in multiple domains, the MetaLWOz corpus. In addition to that, we make use of general-purpose ELMo encoder trained on a large amount of textual data and capturing both character-level and token-level information.

%We evaluate our model on the Stanford Multi-Domain dialogue dataset in terms of BLEU and Entity F1 scores, and show that our approach significantly and consistently improves upon a series of baseline models as well as the previous state-of-the-art dialogue generation model.
%Although the latter is presented as a zero-shot model, we show that our approach requires less in-domain data, with no annotation required.

\section{Introduction}
\label{sec:intro}

Machine learning-based dialogue systems, while still being a relatively new research direction, are experiencing increasingly wide adoption in industry.  Large-scale dialogue assistant platforms such as \textit{Google Assistant, Amazon Alexa, and Apple Siri} provide a unified conversational user interface (CUI) for third-party applications and services. Furthermore, products like \textit{Google Dialogflow, Wit.ai, Microsoft LUIS, and Rasa} offer means for rapid development of a dialogue system's core modules. In addition, with the recently adopted technique of training dialogue systems end-to-end \textit{data-efficiency} of such systems becomes the key question in their adoption in practical applications. Currently, while being extremely flexible and requiring little to no programming of in-domain business logic (see e.g. \citet{DBLP:conf/sigdial/UltesBCRTWYG18, DBLP:conf/icml/WenMBY17, DBLP:conf/eacl/Rojas-BarahonaG17}), such systems have too high data consumption~--- including both collection and annotation effort~--- in order for them to be used in rapidly paced industrial product cycles. Therefore, approaches to training such systems with extremely limited data (i.e.\ zero-, one- and few-shot training) are a priority research direction in the dialogue systems area.

In this paper, we present the \textit{Dialogue Knowledge Transfer Network} (or DiKTNet), a generative goal-oriented dialogue model designed for few-shot learning, i.e.\ training only using a small number of complete in-domain dialogues. The key underlying concept of this model is transfer learning: DiKTNet makes use of the latent text representation learned from several sources ranging from large-scale general-purpose textual corpora to similar dialogues in the domains different to the target one. We use the evaluation framework of \newcite{DBLP:conf/sigdial/ZhaoE18} and the same dataset, and mainly compare our approach to theirs. While their method doesn't require complete in-domain dialogues and uses annotated utterances instead (and is therefore described as ``zero-shot"), we show that our model achieves superior performance with roughly the same amount of data (with respect to in-domain utterances) while requiring no annotations whatsoever.
%\sj{How about saying performance gain in number here?}
% IS: added to abstract

\begin{figure}[ht]
  \centering
  \includegraphics[width=1.0\columnwidth]{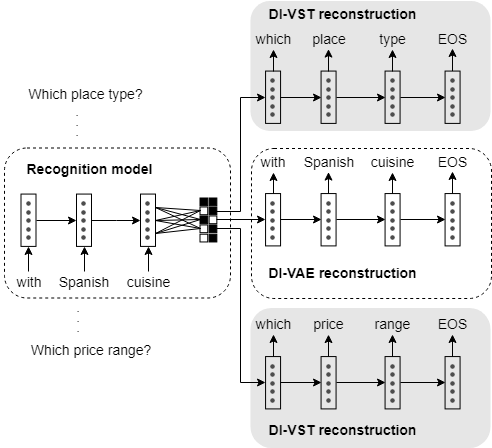}
  \caption{DI-VAE and DI-VST (DiKTNet Stage 1)}
  \label{fig:diktnet_stage1}
\end{figure}

\section{Related Work}
\label{sec:related}

The problem of data efficiency of dialogue systems has been extensively researched in the past. Starting with domain adaptation of a dialogue state tracker \cite{DBLP:conf/slt/HendersonTW14} approached using Bayesian Processes \cite{DBLP:journals/csl/GasicMRSUVWY17} and Recurrent Neural Networks \cite{DBLP:conf/acl/MrksicSTGSVWY15}, there has been significant  work on training different dialogue system components using as little data as possible. As such, \newcite{DBLP:conf/acl/WilliamsAZ17} introduced a dialogue management model designed for bootstrapping from limited training data and further fine-tuning. A recent paper by \newcite{DBLP:journals/corr/abs-1811-11707} introduced a dialogue management model which uses a unified embedding space for user and system turns allowing efficient cross-domain knowledge transfer.

There also exist approaches to end-to-end dialogue generation. \newcite{Eshghi.etal17a} proposed a linguistically informed model based on an incremental semantic parser \cite{Eshghi.etal11} combined with a reinforcement learning-based agent. The parser was used for both maintaining the agent's state and pruning the agent's incremental, word-level generation actions (only the actions leading to syntactically correct word sequences were allowed for the agent to take). While outperforming end-to-end dialogue models on bAbI Dialog Tasks in a zero-shot setup \cite{Shalyminov.etal17} due to its prior linguistic knowledge in the form of a dialogue grammar, this method inherited the limitations of it as well. Specifically, it's limited to a single domain until a wide-coverage grammar is available.

Meta-learning has also gained a lot of attention as a way to train models for maximally efficient adaptation to new data. As such, \newcite{DBLP:conf/acl/QianY19} presented such approach for fast adaptation of a dialogue model to a new domain. While highly promising, its main result was achieved on a synthetic dataset and would ideally need more testing on real data.

Finally, the method we directly compare our approach to is that of \newcite{DBLP:conf/sigdial/ZhaoE18} who introduced the Zero-Shot Dialogue Generation (ZSDG) task and the corresponding model. In their work, they use a unified latent space for user utterances, system turns, and \textit{domain descriptions} in the form of utterance-annotation pairs. Since they only used such utterances and no full dialogues for the target domain, they presented this approach as ``zero-shot" learning. In our approach, we do use complete in-domain dialogues, but with significantly less data with respect to the number of in-domain utterances. Moreover, our method requires no annotation whatsoever.

Recent research in Natural Language Processing has shown that the transfer of text representation learned on larger data sources benefits target models' performance, just as was the case with ImageNet-based computer vision models \cite{imagenet_cvpr09}.

For text, the main means for transfer was Word2Vec and GloVe embeddings \cite{DBLP:conf/nips/MikolovSCCD13,DBLP:conf/emnlp/PenningtonSM14} recently extended with context-aware models like ELMo \cite{DBLP:conf/naacl/PetersNIGCLZ18} and BERT \cite{DBLP:journals/corr/abs-1810-04805}. Trained on large and diverse textual corpora, they were shown to improve target models' performance on a number of Natural Language Processing tasks. Although highly beneficial, those models' use may not be sufficient for the case of dialogue as response generation for goal-oriented dialogue from extremely limited data requires specialized tools. General-purpose embeddings lack specificity for close dialogue domains since they have been learned from very heterogeneously distributed data: in dialogue, the distribution of word sequences is highly specific to a given domain or task, i.e.\ word sequences in dialogue can take on an astonishingly wide variety of meanings in different contexts.

In this paper, we will work with \textit{autoencoders}, a class of unsupervised text representation models working via  reconstructing the input~--- specifically, a Variational Autoencoder (VAE) was considered the main means to learn robust text representations \cite{DBLP:conf/conll/BowmanVVDJB16}. %Although 
%OL why is this sentence staring with "Although"?? there is no contrast made .
However,
the model itself was challenging to train and was mainly used with plenty of workarounds, and recently there started to appear variants of this model with improved stability. One such model we will use in this paper is that of \newcite{DBLP:conf/acl/EskenaziLZ18} (see Section \ref{sec:laed} for more detail).

\section{Few-Shot Dialogue Generation}
\label{sec:fsdg}
We first describe the task we are addressing in this paper, and the corresponding base model.
Specifically, we have a set of dialogues in source domains and just a few seed dialogues in the target domain. The main task of the model is: having been trained on all the available \textit{source data}, to fine-tune on the \textit{target data} to be further evaluated on the full set of target-domain dialogues.

We are basing our model for this task on a Hierarchical Encoder-Decoder (HRED) architecture with attention-based copying \cite{DBLP:conf/iclr/MerityX0S17}. The base optimization objective is as follows:

\begin{equation}
  \label{eq:hred}
  \begin{split}
  & \mathcal{L}_{\text{HRED}} = \log p_{\mathcal{F}^d}(\bm{x}_\text{sys} \mid \mathcal{F}^e(\bm{c}, \bm{x}_\text{usr}))
  \end{split}
\end{equation}

where $\bm{x}_\text{usr}$ is user's query, $\bm{x}_\text{sys}$ is the system's response, $\bm{c}$ is the dialogue context, and $\mathcal{F}^e$ and $\mathcal{F}^d$ are respectively hierarchical encoder and decoder.

We work with goal-oriented dialogues, so it is natural in our setting to take into account an underlying Knowledge Base (or API) providing results for the user's queries.
Given that such KB information may contain unseen token sequences for the most part, especially in the target domain, we use a copy mechanism in order to be able to use this information in the system's responses. More specifically, we represent KB info  as token sequences and concatenate it to the dialogue context similarly to CopyNet setup of \newcite{DBLP:conf/sigdial/EricKCM17}. Our copy mechanism's implementation is the Pointer-Sentinel Mixture Model \cite{DBLP:conf/iclr/MerityX0S17,DBLP:conf/sigdial/ZhaoE18}:

\begin{equation}
\label{eq:psm}
\begin{split}
  p(w_t \mid s_t) = g p_\text{vocab}(w_t \mid s_t) + (1 - g) p_\text{ptr}(w_t \mid s_t)
\end{split}
\end{equation}

In the formula above, $w_t$ and $s_t$ are respectively the output word and the decoder state at step $t$; $p_\text{ptr}$ is the probability of attention-based copying of the word $w_t$, and $g$ is the mixture weight:

\begin{equation}
\label{eq:psm_p_ptr}
\begin{split}
  p_\text{ptr}(w_t \mid s_t) = \sum_{k_j \in I(w, \bm{x})}{\alpha_{k_j, t}}
\end{split}
\end{equation}

\begin{equation}
\label{eq:psm_g}
\begin{split}
  g = \text{Softmax}(u^T \tanh{(W_\alpha s_i)})
\end{split}
\end{equation}

where $\alpha_{k_j, t}$ is the attention weight for $k$th token in flattened dialogue context at the decoding step $t$ and $u$ is the sentinel vector~--- for more detail, see \cite{DBLP:conf/sigdial/ZhaoE18}.

\section{Dialogue Knowledge Transfer Network}
\label{sec:approach}

\begin{figure*}[ht]
  \centering
  \includegraphics[width=1.0\linewidth]{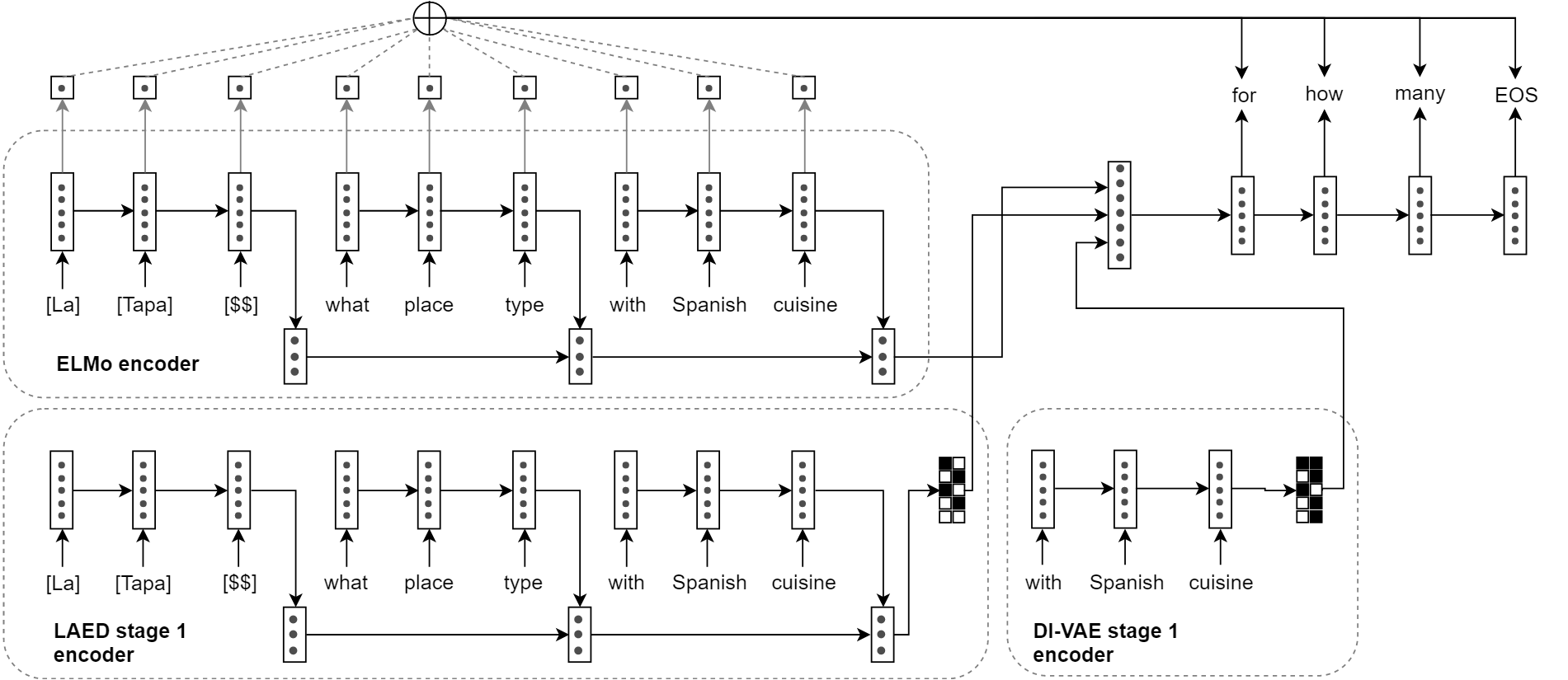}
  \caption{DiKTNet Stage 2 (tokens in brackets are KB data)
  %\sj{Do we run an RNN over DI-VAE and DI-VST vecrtors?}
  %IS: corrected the figure! DI-VST is a full-blown LAED with the hierarchical encoder, DI-VAE is not
  }
  \label{fig:diktnet_stage2}
\end{figure*}

Transfer learning is considered the key means for efficient training with minimal data, and our DiKTNet model essentially introduces several knowledge-transfer augmentations to the base HRED model described above. DiKTNet training is performed in two stages described below.

\subsection{Stage 1. Dialogue representation pre-training}
\label{sec:laed}

Dialogue structure~--- e.g.\ word sequences ~--- is highly specific to a given domain or task, and the meaning of conversational utterances is highly contextual, i.e.\ similar utterances may have different meanings depending on the context. Nevertheless, there is a lot of similarity in dialogue structure~--- i.e.\ sequences of dialogue actions~--- across domains, e.g.\ a conversation normally starts with a mutual greeting and a question is very often followed by an answer.
Here, we propose to exploit this phenomenon in the form of learning a latent dialogue action representation in order to better capture the dialogue structure by abstracting away from surface forms.
% \sj{This is a bit unclear. How about saying capturing such phenomena is not easy due to the variety of surface forms. So, we use latent action representation to better capture the dialogue structure by abstracting away from surface forms.}
% IS: done
Crucially, we learn such representation from MetaLWOz \cite{multi-domain-task-completion-dialog-challenge}, a dataset specifically created for the purposes of meta-learning and transfer learning and consisting of human-human conversations in 51 unique domains (for more detail, see Section \ref{sec:data}).

For this stage of training we use unsupervised, variational autoencoder-based (VAE) representation learning following the \textit{Latent Action Encoder-Decoder (LAED)} approach of \newcite{DBLP:conf/acl/EskenaziLZ18}.
LAED's underlying model is called \textit{Discrete Information VAE (DI-VAE)}, a variant of a VAE with two modifications. Firstly, its optimization objective accounts for the mutual information $I$ between the input and the latent variable which is implicitly discouraged in the original VAE objective (see Eqs.\ \ref{eq:vae_fsdg} and \ref{eq:di_vae}). 

% \sj{Don't forget to refer to the figure.}
% IS: done

\begin{equation}
\label{eq:vae_fsdg}
\begin{split}
  \mathcal{L}_{VAE} & = \mathbb{E}_x[\mathbb{E}_{q_{\mathcal{R}}(\bm{z}\mid\bm{x})} [ \log p_\mathcal{G}(\bm{x} \mid \bm{z})] \\
  & - KL(q(\bm{z}) \rVert p(\bm{z}))] = \\
  & \mathbb{E}_{q(\bm{z} \mid x) p(x)}[ \log p_\mathcal{G}(\bm{x} \mid \bm{z})] \\
  & - I(Z, X) - KL(q(\bm{z}) \rVert p(\bm{z})),
\end{split}
\end{equation}

\begin{equation}
\label{eq:di_vae}
\begin{split}
  \mathcal{L}_{DI\text{-}VAE} & = \mathcal{L}_{VAE} + I (Z, X) \\
  & = \mathbb{E}_{q_{\mathcal{R}}(\bm{z} \mid \bm{x}) p(\bm{x})} [ \log p_\mathcal{G}(\bm{x} \mid \bm{z})] \\
  & - KL(q(\bm{z}) \rVert p(\bm{z}))
\end{split}
\end{equation}

where $\bm{x}$ is the input utterance, $\bm{z}$ is the latent variable ($X$ and $Z$ corresponding to their batch-wise vectors), $\mathcal{R}$ and $\mathcal{G}$ are the recognition and generation models (implemented as RNNs) respectively, and $q(\bm{z}) = \mathbb{E}_x[q_\mathcal{R}(\bm{z} \mid x)]$.

Secondly, the latent variable $\bm{z}$ in DI-VAE is \textit{discrete} as opposed to the continuous one in a vanilla VAE. The discrete latent code lends itself well to interpretation and can be viewed as a form of unsupervised dialogue act tagging.
The discrete nature also makes the calculation of the KL-term more tractable via the \textit{Batch Prior Regularization} technique~\cite{DBLP:conf/acl/EskenaziLZ18}:
% This was used by the authors as a feature to make the system's latent actions interpretable, but they also showed that it makes the calculation of the KL-term more tractable via the \textit{Batch Prior Regularization} technique:

\begin{equation}
\label{eq:bpr}
\begin{split}
  KL(q'(\bm{z}) \rVert p(\bm{z})) = \sum_{k=1}^K{q'(\bm{z} = k) \log \frac{q'(\bm{z} = k)}{p(\bm{z} = k)}}
\end{split}
\end{equation}

where $K$ is the number of $\bm{z}$'s possible values and $q'(\bm{z})$ is the approximation to $q(\bm{z})$ over $N$ data points:

\begin{equation}
\label{eq:q_prime}
\begin{split}
  q'(\bm{z}) = \frac{1}{N}\sum_{n=1}^N{q_{\mathcal{R}}(\bm{z} \mid x_n)}
\end{split}
\end{equation}

In addition, we employ \textbf{\textit{DI-VST}}, DI-VAE's counterpart working in a Variational Skip-Thought manner \cite{DBLP:conf/naacl/HillCK16} and reconstructing the input $\bm{x}$'s previous ($\bm{x_p}$) and next ($\bm{x_n}$) context utterances instead:

\begin{equation}
\label{eq:di_vst}
\begin{split}
  \mathcal{L}_{DI\text{-}VST} & = \\
  & \mathbb{E}_{q_{\mathcal{R}}(\bm{z}\mid\bm{x}) p(\bm{x})} [ \log p^n_\mathcal{G}(\bm{x_n} \mid \bm{z}) p^p_\mathcal{G}(\bm{x_p} \mid \bm{z})] \\
  & - KL(q(\bm{z}) \rVert p(\bm{z}))
\end{split}
\end{equation}

DI-VAE and DI-VST models are visualized in Figure \ref{fig:diktnet_stage1}.

In the downstream DiKTNet model, we use DI-VAE autoencoder in order to obtain the representation of the user's query: $\bm{z}_\text{usr} = \text{DI-VAE}(\bm{x_\text{usr}})$.

In turn, DI-VST is used to obtain a prediction of the system's action $\bm{z}_\text{sys}$ in the discretized latent form given the user's input $\bm{x}_\text{usr}$ as well as the full dialogue context $\bm{c}$. For that, DI-VST autoencoder is used as part of a hierarchical, context-aware encoder-decoder response generation model (we refer to it as \textbf{\textit{LAED}} itself). Its optimization objective is as follows:

\begin{equation}
\label{eq:laed}
\begin{split}
  & \mathcal{L}_{LAED}(\theta_\mathcal{F}, \theta_\pi) = \mathbb{E}_{q_{\mathcal{R}}(\bm{z}_\text{sys}\mid\bm{x}_\text{sys}) p(\bm{x}_\text{sys}, \bm{c})} [ \\
  & \log p_\pi (\bm{z}_\text{sys} \mid \bm{c}) + \log p_\mathcal{F} (\bm{x}_\text{sys} \mid \bm{z}_\text{sys}, \bm{c})]
\end{split}
\end{equation}

%\is{I think there's a typo in Tiancheng's formula: since he denotes system's response as $\bm{\tilde{x}}$, this should be the predicted by the decoder model in the last term}
% \sj{This loss is called LAED in the Tiancheng's paper. So I changed it.}

where $\theta_\mathcal{F}$ is the set of parameters of the context-aware encoder and decoder, $\theta_\pi$ is the set of parameters of the policy $\theta_\pi$. $\theta_\pi$ is the component trained to directly predict $\bm{z}_\text{sys}$ from the context $\bm{c}$.
% \sj{$\mathbb{E}_{p(\bm{x} \mid \bm{c})} [ q_\mathcal{R} (\bm{z} \mid \bm{x})]$ -- this part of TianCheng's paper was confusing and seems incorrect. So I edited the description.}
% IS: OK

We use different models for different aspects of the dialogue: DI-VAE for the user utterance representation, and DI-VST-based LAED  for the system's action prediction. In that, we follow the intuition of \newcite{DBLP:conf/sigdial/ZhaoE18} who argued that DI-VAE is better at capturing specific words of an utterance, while DI-VST represents the overall dialogue action better.

We train these two models on MetaLWOz in an unsupervised way with the objectives as described above, and use their discretized latent codes $\bm{z}_\text{usr}$ and $\bm{z}_\text{sys}$ respectively in the downstream model at the next stage of training.

\begin{table*}[ht!]
  \centering
  \small
    \begin{tabular}{|l||l|l||l|l||l|l|}
      \hline
      \multirow{2}{*}{\backslashbox{\textbf{Model}}{\textbf{Domain}}}&\multicolumn{2}{c||}{\textbf{Navigation}}&\multicolumn{2}{c||}{\textbf{Weather}}&\multicolumn{2}{c|}{\textbf{Schedule}}\\
      &\multicolumn{1}{c|}{BLEU, \%}&\multicolumn{1}{c||}{Entity F1, \%}&\multicolumn{1}{c|}{BLEU, \%}&\multicolumn{1}{c||}{Entity F1, \%}&\multicolumn{1}{c|}{BLEU, \%}&\multicolumn{1}{c|}{Entity F1, \%}\\\hline\hline
      ZSDG&5.9&14.0&8.1&31&7.9&36.9\\
      NLU\_ZSDG&$6.1\pm2.2$&$12.7\pm3.3$&$5.0\pm1.6$&$16.8\pm6.7$&$6.0\pm1.7$&$26.5\pm5.4$\\
      NLU\_ZSDG+LAED&$7.9\pm1$&$12.3\pm2.9$&$8.7\pm0.6$&$21.5\pm6.2$&$8.3\pm1$&$20.7\pm4.8$\\
      % ZSDG&0.061&0.114&0.089&0.062&0.066&0.297\\
      \hline
      HRED@1\%&$6.0\pm1.8$&$9.8\pm4.8$&$6.9\pm1.1$&$22.2\pm 10.7$&$5.5\pm0.8$&$25.6\pm8.2$\\
      HRED@3\%&$7.9\pm0.7$&$11.8\pm4.4$&$9.6\pm1.8$&$39.8\pm 7$&$8.2\pm1.1$&$34.8\pm4.4$\\
      HRED@5\%&$8.3\pm1.3$&$15.3\pm6.3$&$11.5\pm 1.6$&$38.0\pm10.5$&$9.7\pm1.4$&$37.6\pm8.0$\\
      HRED@10\%&$9.8\pm0.8$&$19.2\pm3.2$&$12.9\pm2.4$&$40.4\pm11.0$&$12.0\pm1.0$&$38.2\pm4.2$\\
      \hline
      HRED+VAE@1\%&$3.6\pm2.6$&$9.3\pm4.1$&$6.8\pm1.3$&$23.2\pm10.1$&$4.6\pm1.6$&$28.9\pm7.3$\\
      HRED+VAE@3\%&$6.9\pm1.9$&$15.6\pm5.8$&$9.5\pm2.6$&$32.2\pm11.8$&$6.6\pm1.7$&$34.8\pm7.7$\\
      HRED+VAE@5\%&$7.8\pm1.9$&$12.7\pm4.2$&$10.1\pm2.1$&$40.3\pm10.4$&$8.2\pm1.7$	&$34.2\pm8.7$\\
      HRED+VAE@10\%&$9.0\pm2.0$&$18.0\pm5.8$&$12.9\pm2.2$&$40.1\pm7.6$&$11.6\pm1.5$&	$39.9\pm6.9$\\\hline
      HRED+LAED@1\%&$7.1\pm0.8$&$10.1\pm4.5$&$10.6\pm2.1$&$31.4\pm8.1$&$7.4\pm1.2$&$29.1\pm6.6$\\
      HRED+LAED@3\%&$9.2\pm0.8$&$14.5\pm4.8$&$13.1\pm1.7$&$40.8\pm6.1$&$9.2\pm1.2$&$32.7\pm6.1$\\
      HRED+LAED@5\%&$10.3\pm1.2$&$15.6\pm4.5$&$14.5\pm2.2$&$40.9\pm8.6$&$11.8\pm1.9$&$37.6\pm6.1$\\
      HRED+LAED@10\%&$12.3\pm0.9$&$17.3\pm4.5$&$17.6\pm1.9$&$47.5\pm6.0$&$15.2\pm1.6$	&$38.7\pm8.4$\\\hline
      HRED+ELMo@1\%&$5.8\pm1.9$&$18.2\pm3.8^\mathbf{\star}$&$7.3\pm2.6$&$38.5\pm11.1$&$6.3\pm2.6$&$36.3\pm9.2$\\
      HRED+ELMo@3\%&$8.0\pm1.3$&$17.2\pm4.2$&$10.6\pm1.1$&$42.0\pm11.0$&$9.5\pm2.0$&$39.6\pm9.2$\\
      HRED+ELMo@5\%&$9.4\pm0.8$&$21.5\pm7.3$&$12.1\pm2.0$&$39.0\pm12.8$&$11.3\pm2.1$&$40.0\pm5.6$\\
      HRED+ELMo@10\%&$9.9\pm1.1$&$24.3\pm5.7$&$14.9\pm2.7$&$41.4\pm12.0$&$14.5\pm1.4$&$43.4\pm3.9$\\\hline
      \textbf{DiKTNet@1\%}&$\bm{8.4\pm0.7^*}$&\bm{$15.2\pm4.0}$&$\bm{11.5\pm1.7^*}$&$\bm{43.0\pm10.5^*}$&$\bm{8.1\pm0.8^*}$&$\bm{40.5\pm6.3^*}$\\
      DiKTNet@3\%&$10.4\pm1.2$&$19.2\pm4.8$&$15.7\pm2.1$&$44.0\pm11.7$&$11.1\pm1.3$&$38.2\pm5.8$\\
      DiKTNet@5\%&$11.5\pm1.1$&$23.9\pm2.9$&$15.5\pm2.1$&$39.5\pm14.8$&$13.7\pm2.0$&$41.1\pm3.8$\\
      DiKTNet@10\%&$12.9\pm1.0$&$26.8\pm4.2$&$20.4\pm1.2$&$48.0\pm5.6$&$17.5\pm1.3$&$42.8\pm2.6$\\\hline
    \end{tabular}
    \caption{Evaluation results. Marked with asterisks are individual results higher than ZSDG's performance and which are achieved with the minimum amount of training data. In bold is the model consistently outperforming ZSDG in all domains and metrics with minimum data.}
    \label{tab:results}
\end{table*}

\subsection{Stage 2. Transfer}
At this stage, we train directly for our target task, few-shot dialogue generation, and thus go back to the model described in Section \ref{sec:fsdg}. While the training procedure of this model naturally assumes \textit{domain transfer}, we will provide it with more sources of textual and dialogue knowledge of varying generality described below.

As opposed to direct domain transfer, we incorporate domain-general dialogue understanding from the LAED representation trained on MetaLWOz at the previous stage. 
%MetaLWOz contains dialogues in diverse domains (see Section \ref{sec:data} for more detail), but we make sure to remove any possible overlap with our target data (see Section \ref{sec:setup}), so for our task it can still be considered domain-general. 
LAED captures the background top-down dialogue structure: sequences of dialogue acts in a cooperative conversation, latent dialog act-induced clustering of utterances, and the overall phrase structure of spoken utterances. We incorporate this information into the model by conditioning HRED's decoder on the combined latent codes from Stage 1 and refer to this model as \textit{\textbf{HRED+LAED}}.

\begin{equation}
  \label{eq:hred_laed}
  \begin{split}
  & \mathcal{L}_{\text{HRED+LAED}} = \\
  & \mathbb{E}_{p(\bm{x}_\text{usr},\bm{c}) p (\bm{z}_\text{usr} \mid \bm{x}_\text{usr}) p_\pi (\bm{z}_\text{sys} \mid \bm{x}_\text{usr}, \bm{c})}  [ \\
  & \log p_{\mathcal{F}^d}(\bm{x}_\text{sys} \mid \left \{ \mathcal{F}^e(\bm{x}_\text{usr}, \bm{c}), \bm{z}_\text{usr}, \bm{z}_\text{sys} \right \}) ].
  \end{split}
\end{equation}

where $\bm{z}_\text{usr}$ and $\bm{z}_\text{sys}$ are respectively samples obtained from the DI-VAE user utterance model and LAED/DI-VST system action model, and $ \left \{ \right \}$ is the concatenation operator.

The last, most general source of knowledge we use is a pre-trained ELMo model \cite{DBLP:conf/naacl/PetersNIGCLZ18}. Apart from using an underlying bidirectional RNN encoder, ELMo captures both token-level and character-level information which is especially crucial in understanding unseen tokens and KB items in the underrepresented target domain. The HRED model with ELMo as the utterance-level encoder is referred to as \textit{\textbf{HRED+ELMo}}.

Finally, \textbf{\textit{DiKTNet}} is the HRED augmented with both ELMo encoder and LAED representation.

DiKTNet is visualized in Figure \ref{fig:diktnet_stage2}. The model (as well as its variants listed above) is implemented in PyTorch \cite{paszke2017automatic}, and the code is openly available\footnote{\url{https://bit.ly/fsdg_emnlp2019}}.

% Finally, the last and most specific part of dialogue knowledge comes as domain transfer and is provided by our target dataset, Stanford Multi-Domain dialogue corpus. It contains dialogues in 3 domains: `navigate', `weather', and `schedule' (see more detail in Section \ref{sec:data}). Although the domains differ from each other significantly (e.g. user's goals are expressed as unique sets of slots and values), there are common features shared across all the domains: it is the setting of an intelligent in-car assistant and the use of the underlying knowledge base for the user's queries.

\section{Baselines}

We perform an exhaustive ablation study of DiKTNet by comparing it to all of its variations mentioned above: HRED, HRED+ELMo, and HRED+LAED. In addition to that, we have the \textbf{\textit{HRED+VAE}}~---a version of HRED+LAED for which we use a regular, continuous VAE behind DI-VAE and DI-VST in order to determine the impact of discretized latent codes (see Eq \ref{eq:vae_fsdg} for the corresponding objective function).

Furthermore, we compare DiKTNet to the previous state-of-the-art approach, Zero-Shot Dialogue Generation \cite{DBLP:conf/sigdial/ZhaoE18}. This model didn't use any complete in-domain dialogues but instead it relied on annotated utterances in all of the domains. We use it as-is (\textbf{\textit{ZSDG}}), as well its variation as follows.

We make use of its central idea of `domain descriptions' bridging dialogue understanding across domains, but instead of using manually annotated utterances, we employ automatic Natural Language Understanding markup. Our NLU annotations include:
\begin{itemize}
  \item Named Entity Recognition~--- Stanford NER model ensemble of case-sensitive and caseless models \cite{DBLP:conf/acl/FinkelGM05},
  \item Date/time markup~--- Stanford SUTime \cite{DBLP:conf/lrec/ChangM12},
  \item Wikidata entity linking~--- Yahoo FEL \cite{Blanco:WSDM2015, Pappu:WSDM2017}.
\end{itemize}

We serialize annotations from these sources into token sequences and make domain description tuples out of all the utterances in the source and target domains. In this way, most of our domain descriptions share the structure and content of the original ones.

For example, for the phrase \textit{`Will it be cloudy in Los Angeles on Thursday?'}, the original ZSDG annotation is of the form \texttt{"request \#goal cloudy \#location Los Angeles \#date Thursday"}. Our NLU annotation for this phrase is \texttt{"LOCATION Los Angeles DATE Thursday"}.

We have two models in this setup, with (\textbf{\textit{NLU\_ZSDG+LAED}}) and without the use of LAED representation (\textbf{\textit{NLU\_ZSDG}}) respectively.

\section{Datasets}
\label{sec:data}

\begin{table}[h!]
%\small
\center
\begin{tabular}{ll}
\hline\hline
Number of Domains:&51\\\hline
Number of Dialogues:&40,388\\\hline
Mean dialogue length:&11.91\\
\hline\hline
\end{tabular}
\caption{MetaLWOz dataset statistics}
\label{tab:maluuba}
\end{table}

\begin{table}[h]
\small
\center
\begin{tabular}{|l|l|l|l|}
\hline
\backslashbox{\textbf{Statistic}}{\textbf{Domain}}&Navigation&Weather&Schedule\\\hline\hline
Dialogues&800&797&828\\\hline
Utterances &5248&4314&3170\\\hline
Avg. dialogue length&6.56&5.41&3.83\\\hline
\end{tabular}
\caption{Stanford multi-domain dataset statistics (trainset)}
\label{tab:smd}
\end{table}

We use the Stanford Multi-Domain (SMD) dialogue dataset \cite{DBLP:conf/sigdial/EricKCM17} containing human-human goal-oriented dialogues in three domains: appointment scheduling, city navigation, and weather information. Each dialogue has to do with a single task queried by the user and thus comes with additional knowledge base information coming from implicit querying of the underlying domain-specific API. Although sharing some common features (the setting of an intelligent in-car assistant and the use of the underlying KB), the dialogues differ significantly across domains which makes domain transfer sufficiently challenging.

For the latent representation learning, we use MetaLWOz, a goal-oriented dialogue dataset containing human-human dialogues in diverse domains and several tasks in each of those. The dialogues are collected in a Wizard-of-Oz method where human participants were given a problem domain and a specific task in it, and were asked to complete the task via dialogue. No domain-specific APIs or knowledge bases were available for the participants, and in the actual dialogues they were free to use fictional names and entities in a consistent way. The dataset's statistics are shown in Table \ref{tab:maluuba}. All  domains available in the MetaLWOz dataset are listed in Table \ref{tab:maluuba_domains} of   Appendix A.
% \sj{add example dialogs here} IS: we have plenty of dialogs which is good for readers to get familiar with the dataset, so I think it better stay in the Appendix
%OL double-check that appendix is permitted in submission?

\section{Experimental setup and evaluation}
\label{sec:setup}

\begin{table*}[ht!]
\small
\center
\begin{tabular}{lllll}
\hline\hline
\textbf{Domain}&\multicolumn{2}{c}{\textbf{Context}}&\textbf{Gold response}&\textbf{Predicted response}\\\hline
schedule&\texttt{<usr>}&Remind me to take my pills&Ok setting your medicine&Okay, setting \textit{a reminder to take}\\
&\texttt{<sys>}&What time do you need&appointment for 7pm&\textit{your pills at 7 pm.}\\
&&to take your pills?&&\\
&\texttt{<usr>}&I need to take my pills at 7 pm.&&\\
\hline
navigate&\texttt{<usr>}&Find the address to a hospital&Have a good day&\textit{No problem.}\\
&\texttt{<sys>}&Stanford Express Care is&&\\
&&at 214 El Camino Real.&&\\
&\texttt{<usr>}&Thank you.&&\\
\hline
weather&\texttt{<usr>}&What is the weather forecast&For what city would you&For what city would you like\\
&&for the weekend?&like to know that?&\textit{the weekend forecast for?}\\
\hline\hline
\end{tabular}
\caption{DiKTNet's selected responses}
\label{tab:examples}
\end{table*}

\begin{table*}[t!]
\small
\center
\begin{tabular}{l}
\hline\hline
Where can I go shopping?\\
Where does my friend live?\\
Where can I get Chinese food?\\
Where can I go to eat?\\
Can you please take me to a coffee house?\\
\hline
I'd like to set a reminder for my meeting at 2pm later this month please.\\
What is the time and agenda for my meeting, and who is attending?\\ Schedule a lab appointment with my aunt for the 7th at 1pm.\\
Schedule a calendar reminder for yoga with Jeff at 6pm on the 5th.\\
\hline
Car I'm desiring to do some shopping: which one is it the nearest shopping ...\\ 
... center? Anything within 4 miles?\\
Get the address to my friend's house that i could get to the fastest\\ Car I need to get to my friends house, it should be within 4 miles from here\\
\hline\hline
\end{tabular}
\caption{Selected clusters of utterances sharing the same LAED codes}
\label{tab:laed}
\end{table*}

Our few-shot setup is as follows. Given the target domain, we first train LAED model(s) on the MetaLWOz data~--- and here we exclude from training every domain that might overlap with the target one. Specifically, for the {\it Navigation} domain in SMD, it's {\it Store Details}, for {\it Weather} it's {\it Weather Check}, and for {\it Schedule} it's {\it Update Calendar} and {\it Appointment Reminder}. % {\color{red} AE: this is unclear, are these domains, or actions? Rewrite this.}
% IS: done

In our final setup, at Stage 1 we used a DI-VST-based LAED and a DI-VAE, both of  size $10\times 5$.

Next, having trained and frozen Stage 1 models, we train DiKTNet on all the source domains from the SMD dataset. We use a random sample of the target domain utterances together with their contexts and KB info, varying the amount of those from 1\% to 10\% of all available target data.

For the NLU\_ZSDG setup, we annotated all available SMD data and randomly selected a subset of 1000 utterances from each source domain, and 200 utterances from the target domain. For source domains, this number amounts to roughly a quarter of all available training data~--- we chose it in order to make use of as much annotated data as possible while keeping the domain description task secondary. For the target domain, we made sure to keep under roughly the same in-domain data requirements as the ZSDG baseline.

For evaluation, we follow the approach of \newcite{DBLP:conf/sigdial/ZhaoE18} and report BLEU and Entity F1 scores. Given the non-deterministic nature of our training setup, we report means and variances of our results over 10 runs with different random seeds.

We also perform an additional evaluation of DiKTNet's performance with extended amounts of target data and compare it to the original Key-Value Retrieval Network (\textit{\textbf{KVRet}}) by \newcite{DBLP:conf/sigdial/EricKCM17} which was originally trained with all the available data. In this case we average BLEU scores across all 3 SMD domains in order to be consistent with the form that the corresponding results are presented in the original paper.

We train our models with the Adam optimizer \cite{DBLP:journals/corr/KingmaB14} with learning rate $0.001$. Our hierarchical models' utterance encoder is an LSTM cell \cite{DBLP:journals/neco/HochreiterS97} of   size $256$, and the dialog-level encoder is a GRU \cite{DBLP:conf/emnlp/ChoMGBBSB14} of   size $512$. 

\section{Results and discussion}
\label{ref:results}

Our results are shown in Table \ref{tab:results}~--- our objective here is maximum accuracy with minimum training data required.

\subsection{Results for the few-shot setup}
It can be seen that few-shot models with LAED representation are the best performing models for this objective. While improvements upon ZSDG can already be seen with simple HRED in a few-shot setup, the use of the LAED representation and domain-general ELMo encoding helps significantly reduce the amount of in-domain training data needed: at 1\% of in-domain dialogues, we see that DiKTNet consistently and significantly improves upon ZSDG in every domain. In SMD, with its average dialogue length of 5.25 turns, 1\% of training dialogues amounts to  approximately 40 in-domain training utterances. In contrast, the ZSDG setup used approximately 150 training utterance-annotation pairs for each domain, including the target one, totalling about 450 \textit{annotated} utterances.

Although in our few-shot approach we use full in-domain dialogues, we end up having significantly less in-domain training data, with the crucial difference that none of those has to be annotated for our approach. Therefore, the method we introduced attains state-of-the-art in both accuracy and data-efficiency.

In turn, the results of the  ZSDG\_NLU setup demonstrate that single utterance annotations, if not domain-specific and produced by human experts, don't provide as much signal as full dialogues, even without annotations at all. Even the significant number of such annotated utterances per domain didn't make a difference in this case.

We would also like to point out that, as can be seen in the table, our results have quite high variance~--- the main source of it is the nature of our training/evaluation setup where we average over 10 runs with 10 different sets of seed dialogues. However, in the majority of cases with comparable means, DiKTNet has a lower variance than the alternative models at the same percentage of seed data. And in the extreme case with 1\% target data, DiKTNet improves on all the other models in terms of both means and variances.

\subsection{Discussion of the latent representations}
The comparison of the setups with different latent representations also gives us some insight: while the VAE-powered HRED model improves on the baseline in multiple cases, it lacks generalization potential compared to the LAED setup. The reason for that might be the inherently more stable training of LAED due to its modified objective function, which in turn results in a more informative representation providing better generalization. In order to have a glimpse into the LAED-produced clustering, in Table \ref{tab:laed} we present a snippet of the utterance clusters sharing the same, most frequent latent codes throughout the dataset (the clustering is obtained with LAED model trained on every domain but `Store details', i.e.\ the one for the evaluation on `Navigate' SMD domain). From this snippet, it can be seen that those clusters work well for domain separation, as well as capturing dialogue intents.

\subsection{Results with extended data}
We performed an additional experiment with extended target data (see Figure \ref{fig:extended_data} of Appendix \ref{sec:appendix}). It showed that DiKTNet, when trained with as little as 5\% of target data, can outperform a KVRet trained using the entire dataset. Furthermore, with 50\% of the target data, DiKTNet becomes more than twice as good as KVRet in terms of overall language generation.

However, goal-oriented metrics such as Entity F1 are more challenging to bootstrap. As such, DiKTNet outperforms KVRet on `Weather' domain starting at 10\% of the target data, but only has a trend on narrowing down the performance gap with KVRet on `Navigate', and certainly needs more training data in the `Schedule' domain.

The explanation for  that might be that most of the dialogue entities come from the KB snippets which are the least represented resource in our setup. They aren't available in MetaLWOz, and in SMD, KB snippets share little in common across domains. Therefore, in order to increase Entity F1, KB information should be directly copied to the output more efficiently~--- and increasing the robustness of the copy-augmented decoder is one of our future research directions.

\subsection{Discussion of the evaluation metrics}
We use BLEU as one of the main evaluation metrics in this paper~--- we do it in order to fully conform with the setup of \newcite{DBLP:conf/sigdial/ZhaoE18} which we base our work on. But while being widely adopted as a general-purpose language generation metric, BLEU might not be sufficient in the dialogue setting (see \newcite{Novikova.etal17} for a review). Specifically, we have observed several cases where the model would produce an overall grammatical response with the correct dialogue intent (e.g.\ ``You are welcome! Anything else?''), but BLEU would output a lower score for it due to word mismatch (e.g.\  ``You're welcome!''; see more examples in Table \ref{tab:examples}). This is a general issue in dialogue model evaluation since the variability of possible responses equivalent in meaning is very high in dialogue. In future work, we will put more emphasis on the meaning of utterances, for example by incorporating external dialogue act tagging resources in the evaluation setup which, together with general language generation metrics like perplexity, can make for more robust evaluation criteria than word overlap.
% , using quality metrics of language generation~--- e.g. perplexity~--- as well as keeping task-oriented metrics like Entity F1. We expect these to make for more robust evaluation criteria.

\section{Conclusion and future work}
\label{ref:future}

In this paper, we have introduced DiKTNet, a model achieving state-of-the-art dialogue generation performance in a few-shot setup, without using any annotated data. By transferring latent dialogue knowledge from multiple sources of varying generality, we obtained a model with superior generalization to an underrepresented domain.

Specifically, we showed that our few-shot approach achieves state-of-the art results on the Stanford Multi-Domain dataset while being more data-efficient than the previous best model, by requiring significantly less data none of which has to be annotated.

While being state-of-the-art, the accuracy scores themselves still suggest that our technique is not ready for immediate adoption for real-world production purposes, and the task of few-shot generalization to a completely new dialogue domain remains an area of active research. In our own future work, we will try and find ways to improve the unsupervised representation \cite{DBLP:journals/corr/abs-1904-03736} in order to increase the transfer potential. We will also explore ways to enable more efficient copying from the input which is crucial for correctly handling entities and therefore attaining high goal-oriented performance of the system. 

Apart from that, we will consider alternative evaluation criteria  to account for rich surface variability of natural speech.

\bibliography{all,emnlp-ijcnlp-2019}
\bibliographystyle{acl_natbib}

\newpage
\clearpage
\appendix

\section{Appendices}
\label{sec:appendix}

\begin{table*}[ht]
\small
\center
\begin{tabular}{|ll||ll|}
\hline
\textbf{Domain}&\textbf{\#Dialogues}&\textbf{Domain}&\textbf{\#Dialogues}\\\hline\hline
UPDATE\_CALENDAR&1991&GUINESS\_CHECK&1886\\
ALARM\_SET&1681&SCAM\_LOOKUP&1658\\
PLAY\_TIMES&1601&GAME\_RULES&1590\\
CONTACT\_MANAGER&1483&LIBRARY\_REQUEST&1339\\\hline\hline
INSURANCE&1299&HOME\_BOT&1210\\
HOW\_TO\_BASIC&1086&CITY\_INFO&965\\
TIME\_ZONE&951&TOURISM&935\\
SHOPPING&903&BUS\_SCHEDULE\_BOT&898\\\hline\hline
CHECK\_STATUS&784&WHAT\_IS\_IT&776\\
STORE\_DETAILS&737&APPOINTMENT\_REMINDER&668\\
PRESENT\_IDEAS&664&GEOGRAPHY&653\\
SKI\_BOT&607&MOVIE\_LISTINGS&607\\\hline\hline
UPDATE\_CONTACT&581&ORDER\_PIZZA&577\\
EDIT\_PLAYLIST&574&SPORTS\_INFO&561\\
BOOKING\_FLIGHT&555&WEATHER\_CHECK&551\\
EVENT\_RESERVE&539&RESTAURANT\_PICKER&535\\\hline\hline
LOOK\_UP\_INFO&533&AUTO\_SORT&514\\
QUOTE\_OF\_THE\_DAY\_BOT&513&WEDDING\_PLANNER&510\\
MAKE\_RESTAURANT\_RESERVATIONS&510&AGREEMENT\_BOT&507\\
NAME\_SUGGESTER&499&APARTMENT\_FINDER&499\\\hline\hline
HOTEL\_RESERVE&497&PHONE\_PLAN\_BOT&496\\
DECIDER\_BOT&487&VACATION\_IDEAS&486\\
PHONE\_SETTINGS&473&POLICY\_BOT&447\\
PROMPT\_GENERATOR&446&MUSIC\_SUGGESTER&445\\\hline\hline
PET\_ADVICE&426&BANK\_BOT&367\\
CATALOGUE\_BOT&288&&\\\hline

\end{tabular}
\caption{MetaLWOz dataset domains}
\label{tab:maluuba_domains}
\end{table*}

\subsection{MetaLWOz~--- dialogue examples}
\textbf{Domain}: Ski \\
\textbf{User Task}: You want to know if there are good ski hills an hour's drive from your current location. \\
\textbf{Bot Task}: Tell the user that there are no ski hills in their immediate location. \\
\textcolor{red}{Bot} Hello how may I help you? \\
\textcolor{blue}{User} Is there any good ski hills an hour's drive from my current location? \\
\textcolor{red}{Bot} I'm sorry to inform you that there are no ski hills in your immediate location \\
\textcolor{blue}{User} Can you help me find the nearest? \\
\textcolor{red}{Bot} Absolutely! It looks like you're about 3 hours away from Bear Mountain. That seems to be the closest.\\
\textcolor{blue}{User} Hmm.. sounds good \\
\textcolor{red}{Bot} Alright! I can help you get your lift tickets now!When will you be going? \\
\textcolor{blue}{User} Awesome! please get me a ticket for 10pax \\
\textcolor{red}{Bot} You've got it. Anything else I can help you with? \\
\textcolor{blue}{User} None. Thanks again! \\
\textcolor{red}{Bot} No problem! \\
\\
\textbf{Domain}: Game rules \\
\textbf{User Task}: Ask how to win at the text adventure Zork. If the Bot cannot help you, ask them what games it can help you with.\\
\textbf{Bot Task}: Tell the user that you are programmed to help them play games, not win them. If the user starts talking about other games, go along with them and try to help.  \\ 
\textcolor{red}{Bot} Hello how may I help you? \\
\textcolor{blue}{User} I want to know how I can win playing Zork? \\
\textcolor{red}{Bot} I am programmed to help you play games, not win them \\
\textcolor{blue}{User} What games can you help me with? \\
\textcolor{red}{Bot} Any game \\
\textcolor{blue}{User} Can you specify a certain game? \\
\textcolor{red}{Bot} Zork \\
\textcolor{blue}{User} What are the rules of Zork? \\
\textcolor{red}{Bot} You have to follow through a cave adventure \\
\textcolor{blue}{User} What is the ending of the game? \\
\textcolor{red}{Bot} The user comes out of the cave \\
\\
\textbf{Domain}: Music suggestion \\
\textbf{User Task}: Tell the bot that you like a certain genre of music, but don't prompt it to give you suggestions \\
\textbf{Bot Task}: Provide suggestions, even if the user doesn't ask for them (or tells you they don't need any) \\ 
\textcolor{red}{Bot} Hello how may I help you? \\
\textcolor{blue}{User} Hello.  I like classic rock music. \\
\textcolor{red}{Bot} Dream on Aerosmith \\
\textcolor{blue}{User} No, I'm not a huge fan of that song.  Is there anything else available? \\
\textcolor{red}{Bot} Sweet Home Alabama \\
\textcolor{blue}{User} Well, I do like Lynyrd Skynyrd, but that song is so cliche.  Do you have anything else by the same band? \\ 
\textcolor{red}{Bot} Free bird \\
\textcolor{blue}{User} Yes, that is the perfect song for how I am feeling right now! \\ 
\textcolor{red}{Bot} Great, Is there anything else? \\
\textcolor{blue}{User} No, but I really appreciate your assistance.  Thank you so much! \\
\textcolor{red}{Bot} Youre welcome \\
\\
\textbf{Domain}: Information retrieval \\
\textbf{User Task}: Describe a movie where these guys have to climb a mountain and then throw a thing into the mountain and ask the bot if they know what the movie is called. \\
\textbf{Bot Task}: Ask the user to elaborate as much as possible about the movie they are asking about, and then suggest they're thinking of Lord of the Rings. \\ 
\textcolor{red}{Bot} Hello how may I help you? \\
\textcolor{blue}{User} I'm trying to remember this movie. Can you help? \\
\textcolor{red}{Bot} I would be happy to help.  Can you tell me what it is about? \\
\textcolor{blue}{User} In it these guys have to climb up a mountain. Then they throw something into the mountain. \\
\textcolor{red}{Bot} Do you know what genre the movie is, such as comedy, horror, sci-fi, or fantasy? \\
\textcolor{blue}{User} Not sure about that. \\
\textcolor{red}{Bot} Do you any of the actor or character names in the movie? \\
\textcolor{blue}{User} No... All I can remember is the stuff about the mountain. \\
\textcolor{red}{Bot} Is it a new movie, or an older movie? \\
\textcolor{blue}{User} I don't know. \\
\textcolor{red}{Bot} It sounds like the movie you are talking about could be Lord of the Rings.  Does that sound right? \\
\textcolor{blue}{User} It could be that. Thanks for the help. \\

\begin{figure*}
\centering
\begin{subfigure}[b]{.48\textwidth}
\centering
\pgfplotsset{scaled x ticks=false}
\begin{tikzpicture}[thick,scale=0.78, every node/.style={transform shape}]
  \begin{axis}[
    xlabel={Target data ratio, \%},
    ylabel={BLEU, \% (all domains)},
    xmin=0, xmax=50,
    ymin=0.0, ymax=50,
    xtick={0, 10, 20, 30, 40, 50},
    ytick={0, 10, 20, 30, 40, 50},
    legend style={font=\small},
    legend cell align={right},
    ymajorgrids=true,
    xmajorgrids=true,
    grid style=dotted,
  ]

  \addplot[dashed, color=blue]
  coordinates {
    (0,13.0)(3,13.0)(5,13.0)(10,13.0)(20,13.0)(30,13.0)(40,13.0)(50,13.0)
  };
  \addplot[color=red, mark=square, error bars/.cd, y dir=both, y explicit]
  coordinates {
    (1,9.33) +=(0,1.07) -= (0,1.07)
    (3,12.4) +=(0,1.53) -= (0,1.53)
    (5,13.57) +=(0,1.73) -= (0,1.73)
    (10,16.93) +=(0,1.17) -= (0,1.17)
    (20,18.67) +=(0,1.3) -= (0,1.3)
    (30,20.23) +=(0,1.27) -= (0,1.27)
    (40,21.23) +=(0,1.27) -= (0,1.27)
    (50,21.57) +=(0,1.17) -= (0,1.17)
  };
  \legend{KVRet@100\%, DiKTNet}
  \end{axis}
\end{tikzpicture}
\end{subfigure}
\begin{subfigure}[b]{.48\textwidth}
\centering
\pgfplotsset{scaled x ticks=false}
\begin{tikzpicture}[thick,scale=0.78, every node/.style={transform shape}]
  \begin{axis}[
    xlabel={Target data ratio, \%},
    ylabel={Entity F1, \% (navigate)},
    xmin=0, xmax=50,
    ymin=20, ymax=70,
    xtick={0, 10, 20, 30, 40, 50},
    ytick={0, 10, 20, 30, 40, 50},
    legend style={font=\small},
    legend cell align={right},
    ymajorgrids=true,
    xmajorgrids=true,
    grid style=dotted,
  ]

  \addplot[dashed, color=blue]
  coordinates {
    (0,41.3)(3,41.3)(5,41.3)(10,41.3)(20,41.3)(30,41.3)(40,41.3)(50,41.3)
  };
  \addplot[color=red, mark=square, error bars/.cd, y dir=both, y explicit]
  coordinates {
    (1,15.20) +=(0,4.00) -= (0,4.00)
    (3,19.20) +=(0,4.80) -= (0,4.80)
    (5,23.90) +=(0,2.90) -= (0,2.90)
    (10,26.80) +=(0,4.20) -= (0,4.20)
    (20,29.40) +=(0,3.70) -= (0,3.70)
    (30,29.80) +=(0,3.10) -= (0,3.10)
    (40,29.80) +=(0,3.70) -= (0,3.70)
    (50,32.50) +=(0,3.10) -= (0,3.10)
  };
 
  \legend{KVRet@100\%, DiKTNet}
  \end{axis}
\end{tikzpicture}
\end{subfigure}
\vskip\baselineskip
\begin{subfigure}[b]{.48\textwidth}
\centering
\pgfplotsset{scaled x ticks=false}
\begin{tikzpicture}[thick,scale=0.78, every node/.style={transform shape}]
  \begin{axis}[
    xlabel={Target data ratio, \%},
    ylabel={Entity F1, \% (weather)},
    xmin=0, xmax=50,
    ymin=20, ymax=70,
    xtick={0, 10, 20, 30, 40, 50},
    ytick={0, 10, 20, 30, 40, 50},
    legend style={font=\small},
    legend cell align={right},
    ymajorgrids=true,
    xmajorgrids=true,
    grid style=dotted,
  ]

  \addplot[dashed, color=blue]
  coordinates {
    (0,47)(3,47)(5,47)(10,47)(20,47)(30,47)(40,47)(50,47)
  };
  \addplot[color=red, mark=square, error bars/.cd, y dir=both, y explicit]
  coordinates {
    (1,43.00) +=(0,10.50) -= (0,10.50)
    (3,44.00) +=(0,11.70) -= (0,11.70)
    (5,39.50) +=(0,14.80) -= (0,14.80)
    (10,48.00) +=(0,5.60) -= (0,5.60)
    (20,53.10) +=(0,4.00) -= (0,4.00)
    (30,52.80) +=(0,3.70) -= (0,3.70)
    (40,54.50) +=(0,2.30) -= (0,2.30)
    (50,50.20) +=(0,4.30) -= (0,4.30)
  };
 
  \legend{KVRet@100\%, DiKTNet}
  \end{axis}
\end{tikzpicture}
\end{subfigure}
\begin{subfigure}[b]{.48\textwidth}
\centering
\pgfplotsset{scaled x ticks=false}
\begin{tikzpicture}[thick,scale=0.78, every node/.style={transform shape}]
  \begin{axis}[
    xlabel={Target data ratio, \%},
    ylabel={Entity F1, \% (schedule)},
    xmin=0, xmax=50,
    ymin=20, ymax=70,
    xtick={0, 10, 20, 30, 40, 50},
    ytick={0, 10, 20, 30, 40, 50, 60, 70},
    legend style={font=\small},
    legend cell align={right},
    ymajorgrids=true,
    xmajorgrids=true,
    grid style=dotted,
  ]

  \addplot[dashed, color=blue]
  coordinates {
    (0,62.9)(3,62.9)(5,62.9)(10,62.9)(20,62.9)(30,62.9)(40,62.9)(50,62.9)
  };
  \addplot[color=red, mark=square, error bars/.cd, y dir=both, y explicit]
  coordinates {
    (1,40.50) +=(0,6.30) -= (0,6.30)
    (3,38.20) +=(0,5.80) -= (0,5.80)
    (5,41.10) +=(0,3.80) -= (0,3.80)
    (10,42.80) +=(0,2.60) -= (0,2.60)
    (20,43.00) +=(0,5.30) -= (0,5.30)
    (30,42.00) +=(0,3.80) -= (0,3.80)
    (40,45.10) +=(0,3.20) -= (0,3.20)
    (50,42.80) +=(0,2.80) -= (0,2.80)
  };
 
  \legend{KVRet@100\%, DiKTNet}
  \end{axis}
\end{tikzpicture}
\end{subfigure}
\caption{DiKTNet performance with extended amounts of target data used for training}
\label{fig:extended_data}

\end{figure*}
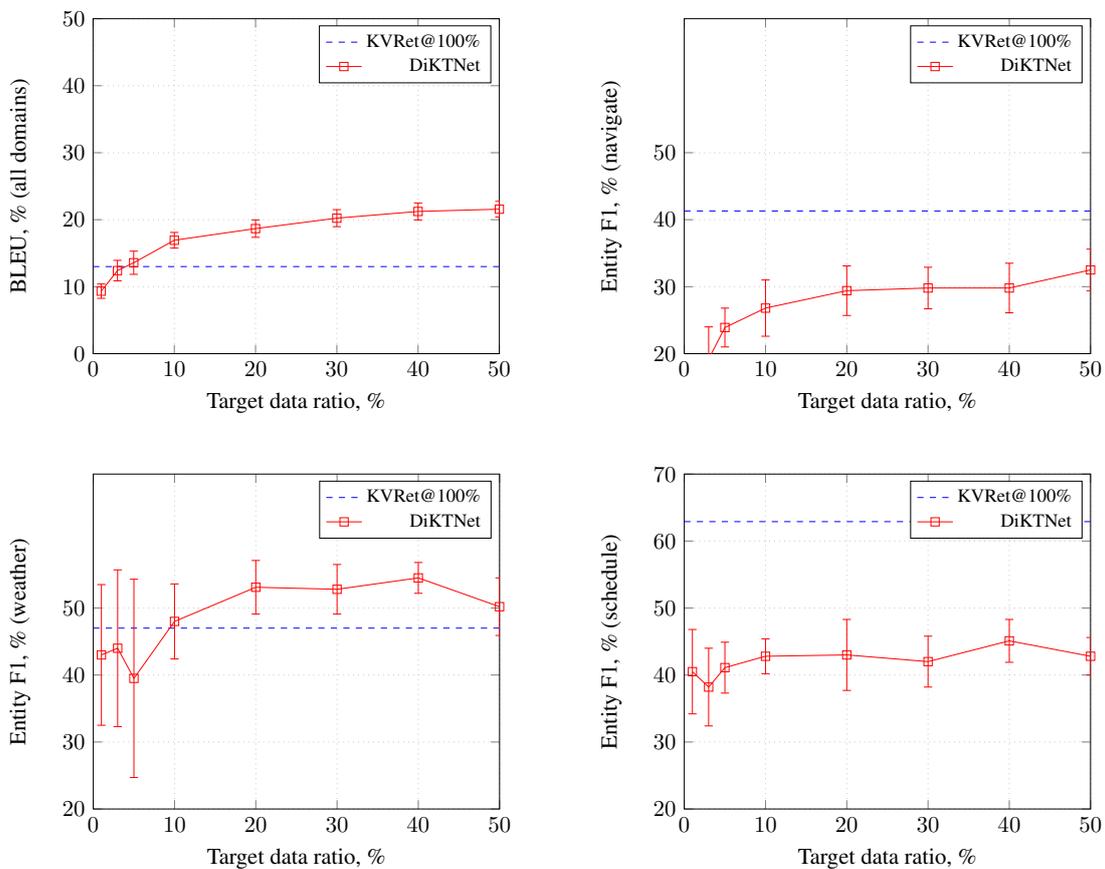
\end{document}